\documentclass{article}
\usepackage{spconf,amsmath,graphicx,algorithm2e}
\usepackage[utf8]{inputenc}

\title{Quality Control in Crowdsourced Object Segmentation}
\twoauthors
{Ferran Cabezas, Axel Carlier, Vincent Charvillat}
{Université de Toulouse\\
	Toulouse, France}
{Amaia Salvador, Xavier Giro-i-Nieto\sthanks{This work has been developed in the framework of the project TEC2013-43935-R, financed by the Spanish Ministerio de Economía y Competitividad and the European Regional Development Fund (ERDF).}}
	{Universitat Politecnica de Catalunya\\
	Barcelona, Catalonia}

\begin{document}
\ninept
\maketitle
\begin{abstract}
This paper explores processing techniques to deal with noisy data in crowdsourced object segmentation tasks. 
We use the data collected with \emph{Click'n'Cut}, an online interactive segmentation tool, and we perform several experiments towards improving the segmentation results.
First, we introduce different superpixel-based techniques to filter users' traces, and assess their impact on the segmentation result. Second, we present different criteria 
to detect and discard the traces from potential bad users, resulting in a remarkable increase in performance. Finally, we show a novel superpixel-based segmentation algorithm which does not require any prior filtering and is 
based on weighting each user's contribution according to his/her level of expertise.

\end{abstract}

\section{Introduction}
\label{sec:intro}

The problem of object segmentation is one of the most challenging ones in computer
vision. It consists in, for a given object in an image, assigning to every pixel
a binary value: 0 if the pixel is not part of the object, and 1 otherwise. Object
segmentation has been extensively studied in various contexts, but still remains
a challenge in general.

In this paper, we focus our experiments on interactive segmentation, that is, object segmentation assisted by human feedback. More 
specifically, we study the particular case in which the interactions come from a
large number of users recruited through a crowdsourcing platform. Relying on humans to help object segmentation is a good idea since the limitations in the semantic interpretation of images is often the bottleneck for computer vision approaches. 

Users, also referred to as \textit{workers} in the crowdsourcing setup, are not experts in the task they must perform and in most cases address it for the first time. 
Workers tend to choose the task that can let them earn the most money in the minimum amount of time. 
From the employer's perspective, crowdsourcing a task to online workers is more affordable than hiring experts. In addition, workers are also available in large numbers and within a short recruiting time.
However, many of these workers are also unreliable and do not meet the minimum quality standards required by the task.
These situations motivate the need for post-processing the collected data to eliminate as many interaction as possible.

\begin{figure}[ht]
  \begin{center}
    \includegraphics[width=0.7\columnwidth]{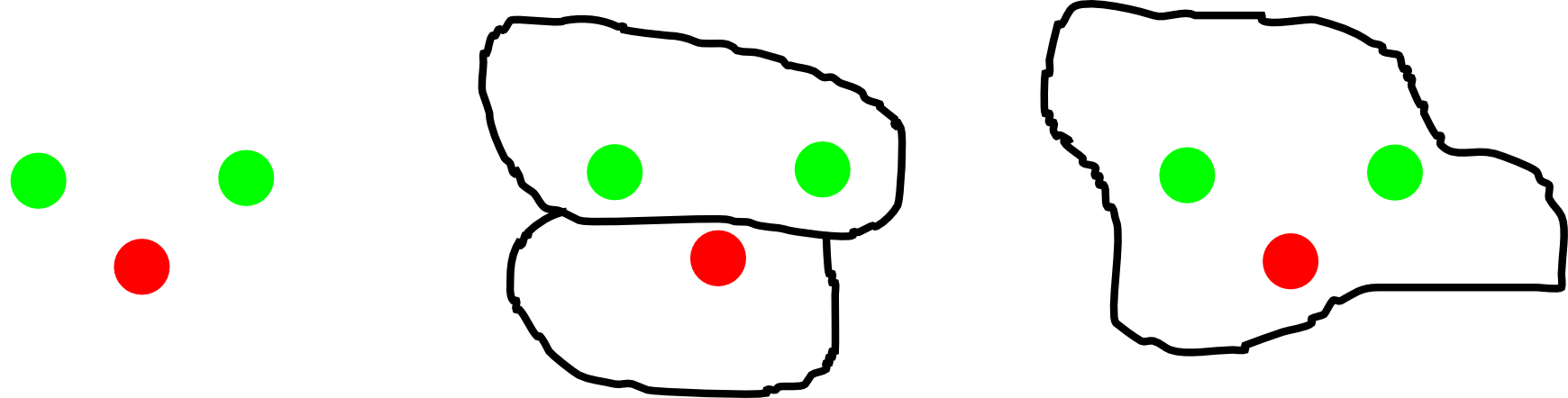}
    \caption{The same set of foreground and background clicks (left) may look consistent (middle) or inconsistent (right) depending on the visual context.}
    \label{fig:intro}
  \end{center}
\end{figure}

Quality control of workers' traces is a very active field of research,  but is also widely dependent on the task. 
In computer vision, the quality of the traces can be estimated with the visual content that motivated their generation. 
As an example, the left side of Figure \ref{fig:intro} depicts 3 points representing the labeling of three pixels: green points for foreground pixels and red points for the background ones. 
These same points may look coherent if assigned to different visual regions (middle) or inconsistent if providing contradictory labels for a the same region (right).
The definition of such regions through an automatic segmentation algorithm can assist in distinguishing between consistent or noisy labels.

This simple example illustrates the assumption that supports this work: computer
vision can help filtering users' inputs as much as users' inputs can guide
computer vision algorithms towards better segmentations. 
Our contributions correspond to the exploration of three different venues for the filtering of human noisy interaction for object segmentation: filtering users, filtering clicks and weighting users' contributions according to a quality estimation.


This paper is structured as follows. Section \ref{sec:relatedwork} overviews previous work in interactive object segmentation and filtering of crowdsourced human traces.
Section \ref{sec:dataacquisition} describes the data acquisition procedure and Section \ref{ssec:context} gives some preliminary results. Then, Section \ref{sec:filtering} introduces the filtering solutions and Section \ref{sec:nonfiltering} explores a user weighted solution.
Finally, Section \ref{sec:conclusion} exposes the conclusions and future work.

\section{Related Work}
\label{sec:relatedwork}


The combination of image processing with human interaction has been extensively explored in the literature. 
Many work related to object segmentation have shown that user inputs throughout a series of weak annotations can be used either to seed segmentation algorithms or to directly produce accurate object segmentations. Researchers have introduced different ways for users to provide annotations for interactive segmentation: by drafting the contour of the objects \cite{Russell-2008-LabelMe, lin2014mscoco}, generating clicks \cite{Carlier-2014-ClicknCut, Salvador-2013-Segmentation, arbelaez2008constrained} or scribbles \cite{rother2004grabcut, McGuinness-InteractiveSegmentation-2010} over foreground and background pixels, or growing regions with the mouse wheel \cite{giro2010gat}.  

However, the performance of all these approaches directly relies on the quality of the traces that users produce, which raises the need for robust techniques to ensure \emph{quality control} of human traces.




The authors in \cite{Oleson-2011-ProgrammaticGold} add gold-standard images in the workflow with a known ground truth to classify users between "scammers", users who do not understand the task and users who just make random mistakes. In \cite{lin2014mscoco}, users are discarded or accepted based on their performance in an initial training task and are periodically verified during the whole annotation process. In any case, authors in \cite{Gottlieb-2012-MechTurk} have demonstrated the need for tutorials by comparing the performance of trained and non trained users. 

Quality control can also be a direct part of the experiment design. 
The \emph{Find-Fix-Verify} design pattern for crowdsourcing experiments was used in \cite{Su-2012-ObjectDetection} for object detection by defining three user roles: a first set of users drew bounding boxes around objects, others verified the quality of the boxes, and a last group checked whether all objects were detected.
Luis Von Ahn also formalized several methods for controlling quality of traces collected from Games With A Purpose (GWAP) \cite{VonAhn-2008-GWAPDesign}. 
Quality control can also be introduced at the end of the study as in \cite{Mao-2013-Incentives}, where a task-specific observation allowed
discarding users whose interaction patterns were unreliable.
Quality control may not be exclusively focused on users but also on the individual traces, as in \cite{Ipeirotis-2010-QualityManagement, welinder2010rating, whitehill2009whose}.
One option to process noisy traces is to collect annotations from different workers and compute a solution by consensus, such as the bounding boxes for object detection computed in \cite{vijayanarasimhan2014active}.



\section{Data Acquisition}
\label{sec:dataacquisition}

The experiment was conducted using the interactive segmentation tool 
\emph{Click'n'Cut} \cite{Carlier-2014-ClicknCut}. 
This tool allows users to label single pixels as  foreground or background, and provides live feedback after each click by displaying the resulting segmentation mask overlaid on the image.

We used the data collected by \cite{Carlier-2014-ClicknCut} over two 
datasets: 
\begin{itemize}
\item 96 images, associated to 100 segmentation tasks, are taken from the DCU dataset \cite{McGuinness-InteractiveSegmentation-2010}, a subset of segmented objects from the Berkeley Segmentation Database \cite{MartinFTM01bsdb}. 
These images will be 
referred in the rest of the paper as our \emph{test set}.
\item 5 images are taken from the PASCAL VOC dataset \cite{Everingham-Pascal-2010}.
We use these images as gold standard, i.e. we use the ground truth of these images to determine workers' errors. These images form our \emph{training set}.
\end{itemize}

Users were recruited on the crowdsourcing platform \emph{microworkers.com}. 
20 users performed the entire set of 105 tasks, 4 females and 16 males, with ages ranging from 20 to 40 (average 25.6). 
Each worker was paid 4 USD when completing the 105 tasks.

\section{Context and previous results}
\label{ssec:context}
The metric we use in this paper is the Jaccard Index, which corresponds to the ratio of the intersection and the union between a segmented object and its ground truth mask, as adopted in the Pascal VOC segmentation task \cite{Everingham-Pascal-2010}.
A Jaccard of 1 is the best possible result (in that case $A = B$), and a Jaccard of
0 means that the two masks have no intersection.


On the test set, experiments on expert users recruited from computer vision research groups reached an average Jaccard of $0.93$ with the best algorithm in \cite{McGuinness-InteractiveSegmentation-2010}. On the other hand, a value $0.89$ was obtained with the same \emph{Click'n'Cut} \cite{Carlier-2014-ClicknCut} tool used in this paper, but on a different group of expert users. However, the group of crowdsourced workers performed significantly worse with \emph{Click'n'Cut}, with a result of $0.14$ with raw traces, which increased up to $0.83$ when filtering worst performing users. In this paper, we propose more sophisticated  filtering techniques to improve this figure.


%
%

\section{Data Filtering}
\label{sec:filtering}

In this section we present three main approaches that focus on filtering
the collected data. Firstly, we present several techniques to filter users' clicks based on their consistency with two image segmentation algorithms. Secondly, we define and apply different rules to discard low quality users. Finally, we explore the combination of both techniques. 

In all the experiments in this section, the filtered data is used to feed the object segmentation algorithm presented in \cite{Carlier-2014-ClicknCut}. This technique generates the object binary mask by combining precomputed \emph{MCG} object candidates \cite{arbelaez2014mcg} according to their correspondence to the users's clicks.

\subsection{Filtering clicks}
\label{sec:filteringclicks}



\begin{figure}[ht]
  \begin{center}
    \includegraphics[width=0.4\columnwidth]{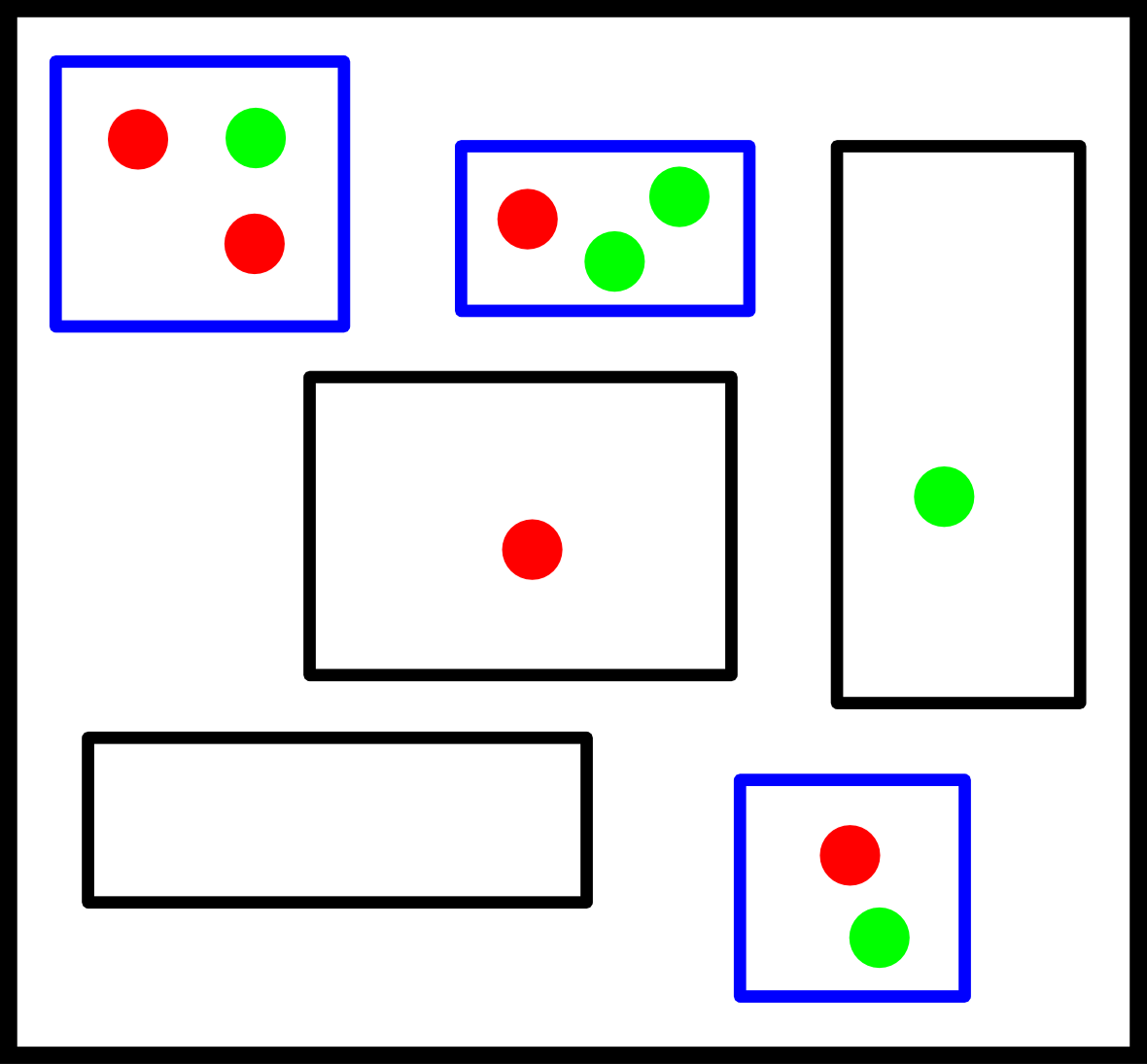}
    \caption[]{Possible configurations of background (in red) and foreground 
(in green) clicks inside a superpixel. Superpixels containing conflicts are represented in blue.}
    \label{fig:filterclicks}
  \end{center}
\end{figure}

Based on the assumption that most of the collected clicks are correct, we postulate that an incorrect click can be detected by looking at other clicks in its spatial neighborhood.
Considering only spatial proximity is not sufficient because the complexity of the object may actually require clicks from different labels to be close, especially near boundaries and salient contours.
For this reason, this filtering relies also on an automatic segmentation of the image, which considers both spatial and visual consistencies.
In particular, image oversegmentations in superpixels have been produced with the SLIC \cite{Achanta-2012-SLIC} and Felzenszwalb \cite{Felzenszwalb2004} algorithms. 
Figure \ref{fig:filterclicks} shows the 6 possible click distributions that can occur in a given superpixel (as shown in 
figure \ref{fig:filterclicks}): 
higher number of foreground than background clicks, 
higher number of background than foreground clicks,
same number of background and foreground clicks,
foreground clicks only,
background clicks only
and no clicks.


\begin{figure}[ht]
  \begin{center}
    \includegraphics[width=0.4\columnwidth]{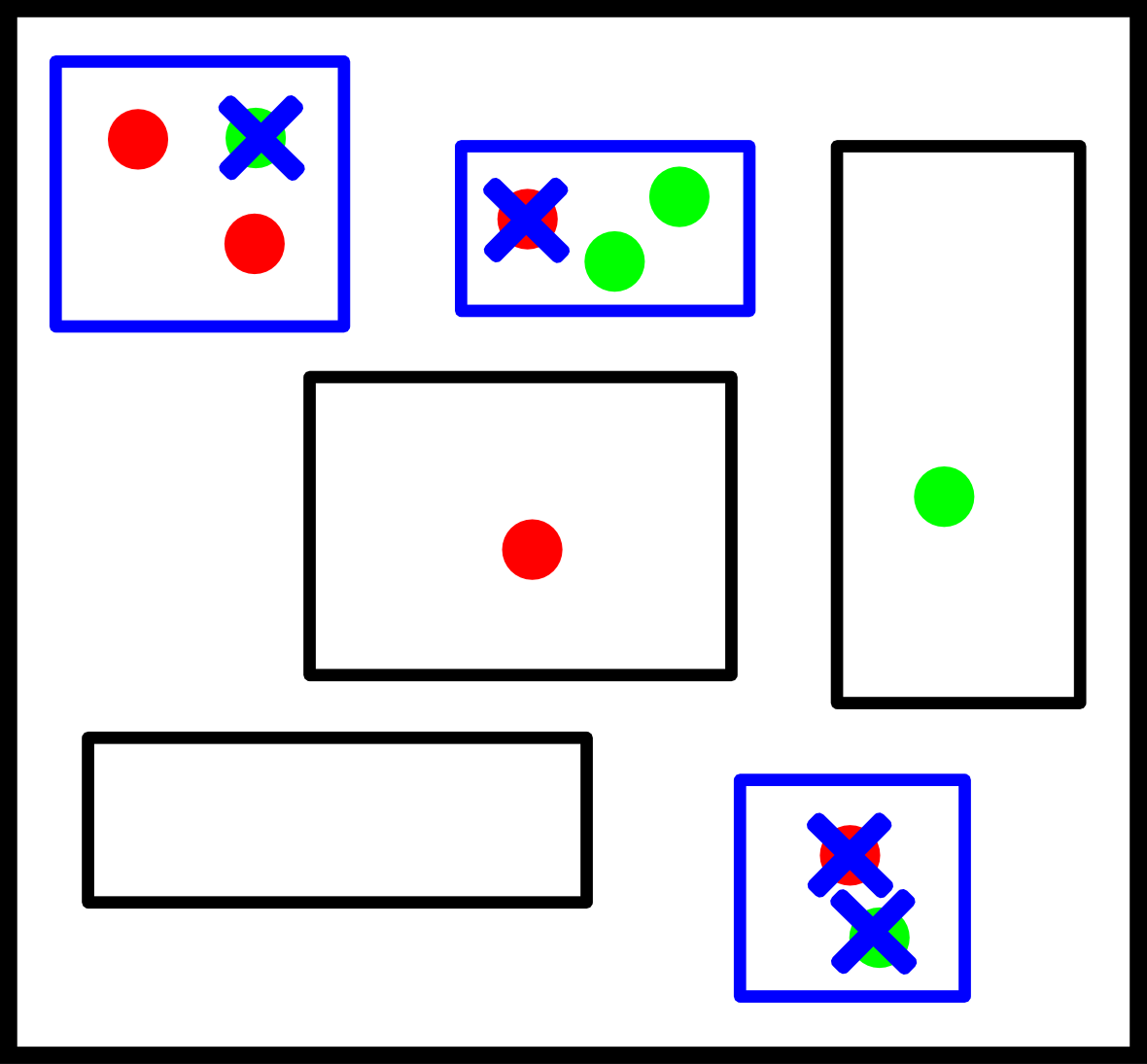}
    \includegraphics[width=0.4\columnwidth]{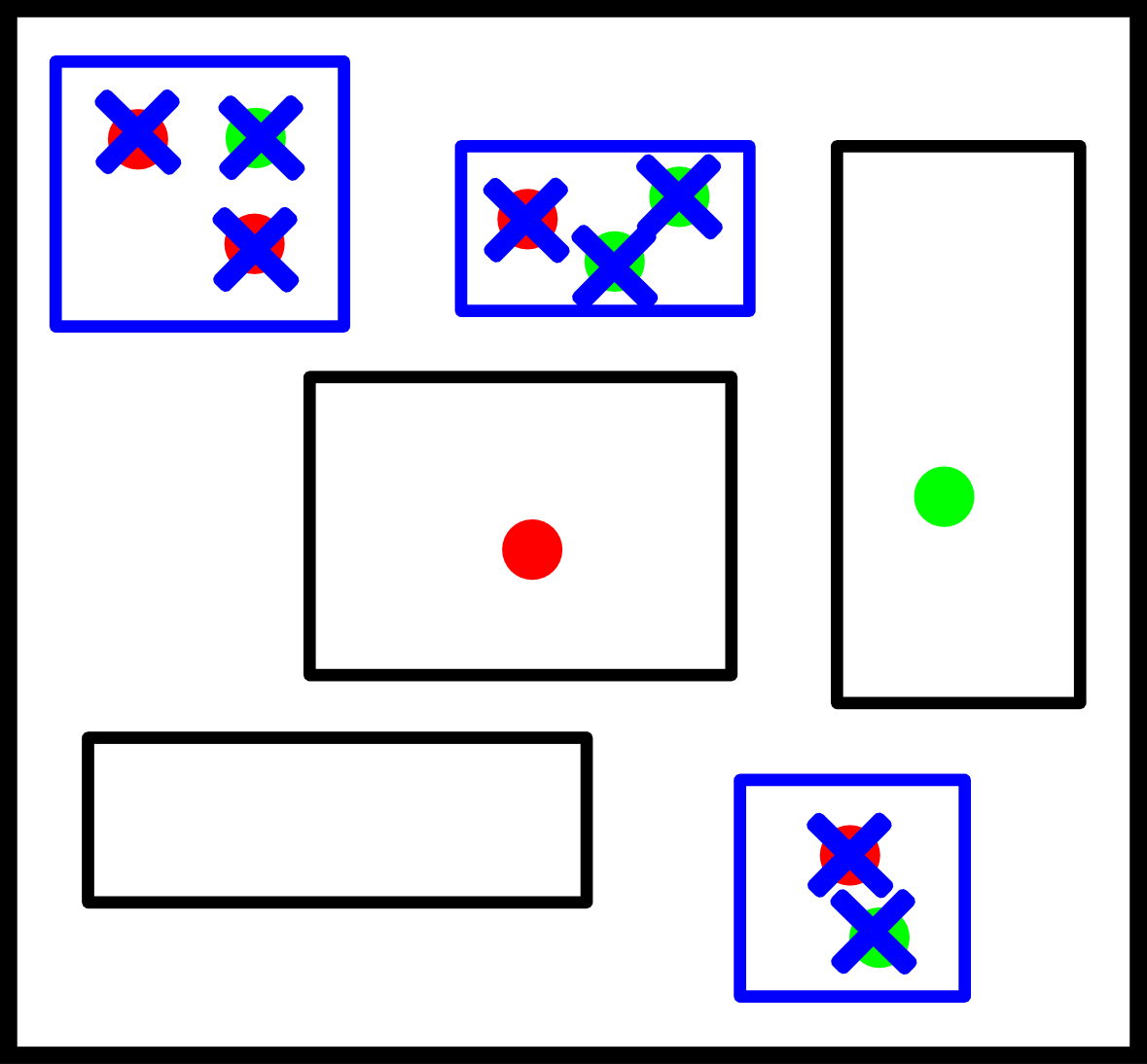}
    \caption[]{Two options to solve conflicts: keep majorities (on the left) and discard all (right).}
    \label{fig:filterclickstechniques}
  \end{center}
\end{figure}

Among these six configurations, the three first ones reveal conflicts between clicks. 
Figure \ref{fig:filterclickstechniques} depicts the two different methods that have been considered to solve the conflicts: keep only those clicks which are majority within the superpixel (left), or discard all conflicting clicks (right).


\begin{table}[h]
\centering
\begin{tabular}{|c|c|c|}
\hline
	& Keep majority 	& Discard all  \\
\hline
SLIC \cite{Achanta-2012-SLIC} & 0.21 (+50\%) & 0.24(+71.43\%)  \\
\hline
Felzenszwalb \cite{Felzenszwalb2004} & 0.21 (+50\%) & 0.22 (+57.14\%) \\
\hline
\end{tabular}
\caption[]{Jaccard Index obtained on the test set after applying the two proposed filtering techniques on \cite{Achanta-2012-SLIC} or \cite{Felzenszwalb2004} superpixels. The Jaccard without filtering is equal to 0.14, so the percentage values in parentheses correspond to the gain with respect to this baseline.}
\label{tab:jaccfilter} 
\end{table}

Table \ref{tab:jaccfilter} shows a significant gain by filtering clicks based on superpixels. However, Jaccard indexes are still too low to consider segmentations useful. Further sections explore other solutions that take into consideration quality control of users in addition to label coherence within superpixels.


\subsection{Filtering users}
\label{sec:filteringusers}

In any crowdsourcing task, recruiting low quality workers is the norm, not the exception. 
In this section we propose to use our training set as a gold standard to determine which users should be ignored.
In particular, two features are computed to decide between accepted and rejected users: their click error rate and their average Jaccard index.



\begin{figure}[ht]
  \begin{center}
    \includegraphics[width=\columnwidth]{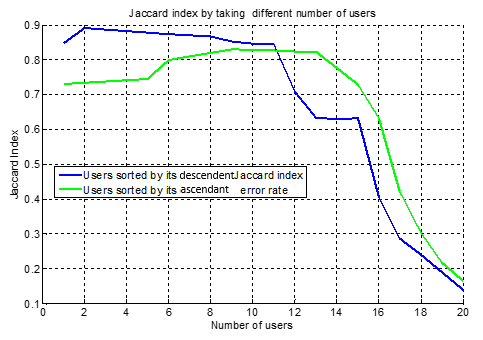}
    \caption[]{Jaccard index (Y-axis) obtained when considering only the top $N$ users (X-axis) according to their average Jaccard (blue) or labeling error rate (green).}
    \label{fig:jaccvserrfilter}
  \end{center}
\end{figure}

Figure \ref{fig:jaccvserrfilter} plots two graphs depicting the average Jaccard by keeping the top $N$ users according to their click error rate or personal Jaccard index. The main conclusion that can be derived from this graph is that personal Jaccard performs better than click errror rate to estimate the quality of the workers. The error rate is not discriminant enough to filter out some types of users: spammers do not necessarily make a lot of mistakes, users who do not understand the task may still produce valid clicks, and good users may also get tired and produce errors on a few images.
For this reasons, it seems more effective to filter users based on their actual performance on the final task (i.e. Jaccard Index for the problem of object segmentation) than in some intermediate metric.




The Jaccard-based curve (blue) from Figure \ref{fig:jaccvserrfilter} shows how the best result is achieved when considering only the two best workers, with a Jaccard of $0.9$ comparable to what expert users had reached (see Section \ref{ssec:context}).
It could be argued that two users are not significant enough and that reaching such a high value as $0.9$ could be a statistical anomaly.
Nevertheless, if many more users are considered and clicks from the top half users are processed, a still high Jaccard of nearly $0.85$ is achieved. This result indicates that filtering users has a much greater impact than just filtering clicks, as presented in Section \ref{sec:filteringclicks}, where the best Jaccard obtained was $0.24$.

\subsection{Filtering clicks and users}
\label{sec:filteringusersandclicks}

This section explores whether, once users have been filtered as explained in Section \ref{sec:filteringusers}, the click-based filtering presented in Section \ref{sec:filteringclicks} can further clean the remaining set of clicks. 

\begin{figure}[ht]
  \begin{center}
    \includegraphics[width=0.8\columnwidth]{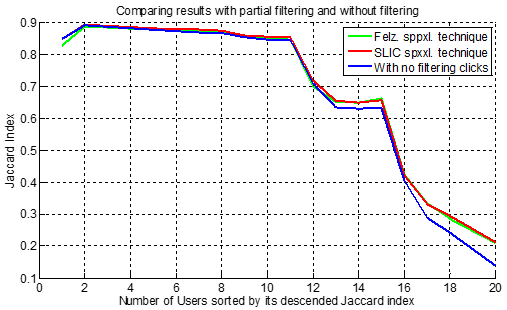}
    \caption[]{Segmentation results with the best $N$ users according to their personal Jaccard-based quality estimation. Red and green curves consider  filtering by majority, while blue curve does not apply any click filtering.}
    \label{fig:userfilter+partial}
  \end{center}
\end{figure}

Figure \ref{fig:userfilter+partial} shows the Jaccard curves obtained when 
applying the majority-based filtering after user filtering. Graphs indicate that there is no major effect when considering a low number of higher quality users, but that the effect is more significant when adding worse users.

\begin{figure}[ht]
  \begin{center}
    \includegraphics[width=0.8\columnwidth]{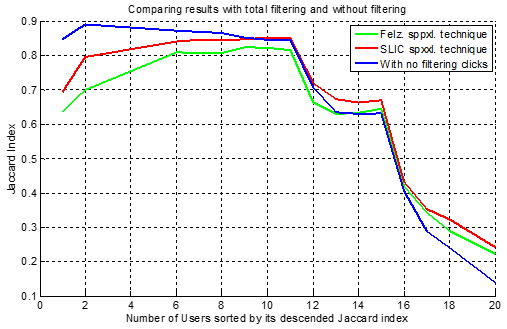}
    \caption[]{Segmentation results with the best $N$ users according to their personal Jaccard-based quality estimation. Red and green curves discard all conflicting clicks, while blue curve does not apply any click filtering.}
    \label{fig:userfilter+total}
  \end{center}
\end{figure}

The case of filtering all conflicting clicks is studied in Figure \ref{fig:userfilter+total}. In this situation, this filtering 
causes a severe drop in performance when few users are considered, and has 
mostly the same effect as majority filtering otherwise. 
This is probably explained by the fact that discarding all clicks when few users are considered results too aggressive and does not provide enough labels to choose a good combination of object candidates.

\section{Data weighting}
\label{sec:nonfiltering}

In section \ref{sec:filtering} we have presented how removing some of the collected user clicks could improve the segmentation results. 
Unfortunately, adopting hard decision criteria may sometimes result into also discarding clicks which may be correct and useful when analyzed as part of a more global problem.
This is why we propose in this section a softer approach that combines the entire set of clicks without any filtering.




The first difference with Section \ref{sec:filtering} is that users are not simply accepted or rejected, but their contribution is weighted according to an estimation of their quality.
A quality score $q_i$ is computed for each user $i$ based on their traces on the gold standard images (see Section \ref{sec:filteringusers} for details). 
The second difference with respect to Section \ref{sec:filtering} is that instead of using object candidates, this time superpixels are used to directly determine the object boundaries.
In particular, the two same segmentation algorithms used in Section \ref{sec:filtering} (Felzenswalb \cite{Felzenszwalb2004} and SLIC \cite{Achanta-2012-SLIC}), are adopted to generate multiple oversegmentations over the image.
In particular, a first set of image partitions were generated by running the technique from Felzenswalb \cite{Felzenszwalb2004} with its parameter $k$ equal to 10, 20, 50, 100, 200, 300, 400 and 500; and a second set of partitions generated with SLIC  \cite{Achanta-2012-SLIC} considering as initial region size 5, 10, 20, 30, 40 and 50 pixels. 
These combinations of parameters were determined after experimentation on the training set.
User clicks with quality estimation and the set of partitions were fed into Algorithm \ref{al:foreground} to generate a binary mask for each object.

\begin{algorithm}[htbp]
\KwData{clicks from all users with their quality scores}
 \KwData{set of segmentations computed from the image}
 \KwResult{binary mask of the segmented object}
 initialize all superpixel scores to 0\;
 
 \While{not all segmentations are processed}{
 	read current segmentation\;
    
 	\While{not all users are processed}{
		read quality estimation $q_j$ from current user $j$\;
        
   		\While{not all clicks from current user $j$ are read}{
       
    		read current click from user $j$\;            	
    		read superpixel corresponding to the click\;
    		
            \eIf{click label is foreground}{
     			add $q_j$ to the current superpixel score\;
     		}{
     		add $1-q_j$ to the current superpixel score\;
            }
    	}
  	}
    compute the average score for each superpixel\;
    normalize superpixels values between 0 and 1\;
 }
 average weighted segmentations to obtain a foreground map\;
 binarize foreground map to obtain the object mask\;

 \caption{Computation of the foreground map}
 \label{al:foreground}
\end{algorithm}




 
\begin{figure}[ht]
  \begin{center}
    \includegraphics[width=0.48\columnwidth]{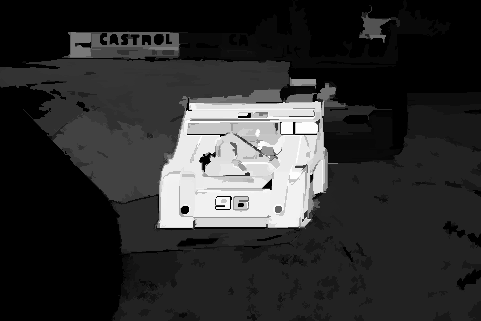}
    \includegraphics[width=0.48\columnwidth]{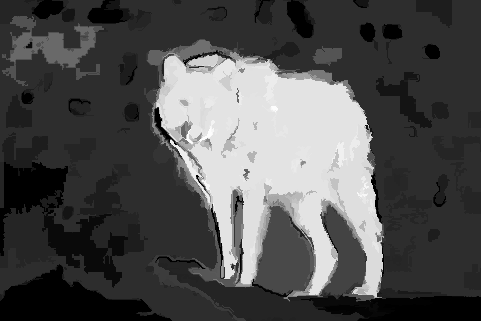}
    \caption[]{Foreground map of object segmentation based on weighted worker's clicks.}
    \label{fig:felz}
  \end{center}
\end{figure}

Figure \ref{fig:felz} gives two examples of foreground maps, with images that contain values ranging from 0 (maximum confidence of \textit{background}) to 1 (maximum confidence of \textit{foreground}).
The object to be segmented is the brightest region, and traces from noisy clicks can be 
seen where regions in the background are bright as well.
As indicated in the last step of Algorithm \ref{al:foreground}, the object masks were obtained by binarizing the foreground maps by applying a threshold equal to $0.56$, also learned on the training set. 
As a final result, this configuration produced a an averaged Jaccard index equal to $0.86$.

\section{Conclusion}
\label{sec:conclusion}

This work has explored error resilience strategies for the problem of object segmentation in crowdsourcing.
Two main directions were addressed: a hard filtering of users and clicks based on superpixels, and a softer solution based on the quality estimation of users and combination of multiple image partitions.

The proposed strategies for filtering clicks based on superpixel coherence introduced significant gains with respect to previous works, but the final quality was still too low.
Our experiments indicate that more significant gains can be obtained by estimating the quality of
each individual user on gold standard tasks. We also show that estimating users quality based on their
performance in the segmentation task is more reasonable than just based on the error rate of the clicks they generate.
Our data indicates that identifying very few high quality workers can produce really high results (0.9 with top two users), even better than the results of expert users with with the same platform (0.89) \cite{Carlier-2014-ClicknCut} and comparable to results of other expert users using different tools \cite{McGuinness-InteractiveSegmentation-2010} (0.93). 

Assuming that very high quality users will always be available in a crowdsourcing campaign may be 
too restrictive.
As an alternative, considering all data with a soft weighting approach seems a more robust
approach compared to the hard filtering and selection of object candidates. 
Our algorithm that weights superpixels according to crowdsourcing clicks (Section \ref{sec:nonfiltering}) has
achieved a significant Jaccard Index of 0.86 without discarding any users or clicks. 
In addition, we have observed that combining the superpixels of multiple sizes and from two different segmentation algorithms (SLIC and Felzenszwalb) seems complementary and benefits the results.

The presented results indicate the potential of using image processing algorithms for quality control of noisy human interaction, also when such interaction may eventually be used to train computer vision systems. In fact, it is the combination of the crowd (majority of correct clicks) and image processing (superpixels) which allows the detection and reduction of a minority of noisy interactions.



\bibliographystyle{IEEEbib}
\bibliography{refs}

\begin{thebibliography}{10}

\bibitem{Russell-2008-LabelMe}
Bryan~C Russell, Antonio Torralba, Kevin~P Murphy, and William~T Freeman,
\newblock ``Labelme: a database and web-based tool for image annotation,''
\newblock {\em International journal of computer vision}, vol. 77, no. 1-3, pp.
  157--173, 2008.

\bibitem{lin2014mscoco}
Tsung-Yi Lin, Michael Maire, Serge Belongie, James Hays, Pietro Perona, Deva
  Ramanan, Piotr Doll{\'a}r, and C.~Lawrence Zitnick,
\newblock ``Microsoft coco: Common objects in context,''
\newblock {\em CoRR}, 2014.

\bibitem{Carlier-2014-ClicknCut}
Axel Carlier, Vincent Charvillat, Amaia Salvador, Xavier Giro-i Nieto, and Oge
  Marques,
\newblock ``Click'n'cut: crowdsourced interactive segmentation with object
  candidates,''
\newblock in {\em Proceedings of the 2014 International ACM Workshop on
  Crowdsourcing for Multimedia}. ACM, 2014, pp. 53--56.

\bibitem{Salvador-2013-Segmentation}
Amaia Salvador, Axel Carlier, Xavier Giro-i Nieto, Oge Marques, and Vincent
  Charvillat,
\newblock ``Crowdsourced object segmentation with a game,''
\newblock in {\em Proceedings of the 2nd ACM international workshop on
  Crowdsourcing for multimedia}. ACM, 2013, pp. 15--20.

\bibitem{arbelaez2008constrained}
Pablo Arbel{\'a}ez and Laurent Cohen,
\newblock ``Constrained image segmentation from hierarchical boundaries,''
\newblock in {\em Computer Vision and Pattern Recognition, 2008. CVPR 2008.
  IEEE Conference on}. IEEE, 2008, pp. 1--8.

\bibitem{rother2004grabcut}
Carsten Rother, Vladimir Kolmogorov, and Andrew Blake,
\newblock ``Grabcut: Interactive foreground extraction using iterated graph
  cuts,''
\newblock in {\em ACM Transactions on Graphics (TOG)}. ACM, 2004, vol.~23, pp.
  309--314.

\bibitem{McGuinness-InteractiveSegmentation-2010}
Kevin McGuinness and Noel~E. O'Connor,
\newblock ``A comparative evaluation of interactive segmentation algorithms,''
\newblock {\em Pattern Recognition}, vol. 43, no. 2, 2010.

\bibitem{giro2010gat}
Xavier Giro-i Nieto, Neus Camps, and Ferran Marques,
\newblock ``Gat: a graphical annotation tool for semantic regions,''
\newblock {\em Multimedia Tools and Applications}, vol. 46, no. 2-3, pp.
  155--174, 2010.

\bibitem{Oleson-2011-ProgrammaticGold}
David Oleson, Alexander Sorokin, Greg~P Laughlin, Vaughn Hester, John Le, and
  Lukas Biewald,
\newblock ``Programmatic gold: Targeted and scalable quality assurance in
  crowdsourcing.,''
\newblock {\em Human computation}, vol. 11, pp. 11, 2011.

\bibitem{Gottlieb-2012-MechTurk}
Luke Gottlieb, Jaeyoung Choi, Pascal Kelm, Thomas Sikora, and Gerald Friedland,
\newblock ``Pushing the limits of mechanical turk: qualifying the crowd for
  video geo-location,''
\newblock in {\em Proceedings of the ACM multimedia 2012 workshop on
  Crowdsourcing for multimedia}. ACM, 2012, pp. 23--28.

\bibitem{Su-2012-ObjectDetection}
Hao Su, Jia Deng, and Li~Fei-Fei,
\newblock ``Crowdsourcing annotations for visual object detection,''
\newblock in {\em Workshops at the Twenty-Sixth AAAI Conference on Artificial
  Intelligence}, 2012.

\bibitem{VonAhn-2008-GWAPDesign}
Luis Von~Ahn and Laura Dabbish,
\newblock ``Designing games with a purpose,''
\newblock {\em Communications of the ACM}, vol. 51, no. 8, pp. 58--67, 2008.

\bibitem{Mao-2013-Incentives}
Andrew Mao, Ece Kamar, Yiling Chen, Eric Horvitz, Megan~E Schwamb, Chris~J
  Lintott, and Arfon~M Smith,
\newblock ``Volunteering versus work for pay: Incentives and tradeoffs in
  crowdsourcing,''
\newblock in {\em First AAAI Conference on Human Computation and
  Crowdsourcing}, 2013.

\bibitem{Ipeirotis-2010-QualityManagement}
Panagiotis~G Ipeirotis, Foster Provost, and Jing Wang,
\newblock ``Quality management on amazon mechanical turk,''
\newblock in {\em Proceedings of the ACM SIGKDD workshop on human computation}.
  ACM, 2010, pp. 64--67.

\bibitem{welinder2010rating}
P.~Welinder and P.~Perona,
\newblock ``Online crowdsourcing: Rating annotators and obtaining
  cost-effective labels,''
\newblock in {\em Computer Vision and Pattern Recognition Workshops (CVPRW),
  2010 IEEE Computer Society Conference on}, June 2010, pp. 25--32.

\bibitem{whitehill2009whose}
Jacob Whitehill, Ting-fan Wu, Jacob Bergsma, Javier~R Movellan, and Paul~L
  Ruvolo,
\newblock ``Whose vote should count more: Optimal integration of labels from
  labelers of unknown expertise,''
\newblock in {\em Advances in neural information processing systems}, 2009, pp.
  2035--2043.

\bibitem{vijayanarasimhan2014active}
Sudheendra Vijayanarasimhan and Kristen Grauman,
\newblock ``Large-scale live active learning: Training object detectors with
  crawled data and crowds,''
\newblock {\em International Journal of Computer Vision}, vol. 108, no. 1-2,
  pp. 97--114, 2014.

\bibitem{MartinFTM01bsdb}
D.~Martin, C.~Fowlkes, D.~Tal, and J.~Malik,
\newblock ``A database of human segmented natural images and its application to
  evaluating segmentation algorithms and measuring ecological statistics,''
\newblock in {\em Proc. 8th Int'l Conf. Computer Vision}, July 2001, vol.~2,
  pp. 416--423.

\bibitem{Everingham-Pascal-2010}
M.~Everingham, L.~Van~Gool, C.~K.~I. Williams, J.~Winn, and A.~Zisserman,
\newblock ``The pascal visual object classes (voc) challenge,''
\newblock {\em IJCV}, vol. 88, no. 2, 2010.

\bibitem{arbelaez2014mcg}
P.~Arbelaez, J.~Pont-Tuset, J.~Barron, F.~Marques, and J.~Malik,
\newblock ``Multiscale combinatorial grouping,''
\newblock in {\em Computer Vision and Pattern Recognition (CVPR), 2014 IEEE
  Conference on}, June 2014, pp. 328--335.

\bibitem{Achanta-2012-SLIC}
Radhakrishna Achanta, Appu Shaji, Kevin Smith, Aurelien Lucchi, Pascal Fua, and
  Sabine Susstrunk,
\newblock ``Slic superpixels compared to state-of-the-art superpixel methods,''
\newblock {\em Pattern Analysis and Machine Intelligence, IEEE Transactions
  on}, vol. 34, no. 11, pp. 2274--2282, 2012.

\bibitem{Felzenszwalb2004}
Pedro Felzenszwalb and Daniel Huttenlocher,
\newblock ``Efficient graph-based image segmentation,''
\newblock {\em International Journal of Computer Vision}, vol. 59, no. 2, pp.
  167--181, 2004.

\end{thebibliography}

\end{document}